\documentclass{article}
\usepackage[preprint]{spconf}
\usepackage{amsmath,graphicx}

\usepackage{amssymb}
\usepackage[dvipsnames,table,xcdraw]{xcolor}
\usepackage{booktabs}
\usepackage{siunitx}
\usepackage{tabularx}
\usepackage{multirow}
\usepackage{caption}
\usepackage{xspace}
\usepackage{arydshln}
\usepackage{subcaption}
\usepackage{hyperref}

\newcommand{\NETNAME} {GANzzle\xspace} 

                \toappear{\copyright\ IEEE 2022. Published in 2022 IEEE International Conference on Image Processing (ICIP), Oct 16-19, Bordeaux, France.}

\title{GANzzle: Reframing jigsaw puzzle solving as a retrieval task using a generative mental image}

\name{Davide Talon$^{\dagger,\ddagger}$, Alessio {Del Bue}$^{\dagger}$, Stuart James$^{\dagger}$ 
\thanks{This project has received funding from the European Union’s Horizon 2020 research and innovation programme under grant agreement No 964854.}
}
\address{$^{\dagger}$Pattern Analysis and Computer Vision (PAVIS), Istituto Italiano di Tecnologia (IIT), Italy \\
$^{\ddagger}$Universit\`{a} degli studi di Genova, Italy}
\begin{document}
%
\maketitle
\begin{abstract}
Puzzle solving is a combinatorial challenge due to the difficulty of matching adjacent pieces. %
Instead, we infer a mental image from all pieces, which a given piece can then be matched against avoiding the combinatorial explosion. Exploiting advancements in Generative Adversarial methods, we learn how to reconstruct the image given a set of unordered pieces, allowing the model to learn a joint embedding space to match an encoding of each piece to the cropped layer of the generator. Therefore we frame the problem as a $R@1$ retrieval task, and then solve the linear assignment using differentiable Hungarian attention, making the process end-to-end. 
In doing so our model is puzzle size agnostic, in contrast to prior deep learning methods which are single size.
We evaluate on two new large-scale datasets, where our model is on par with deep learning methods, while generalizing to multiple puzzle sizes.  
\end{abstract}
\begin{keywords}
Jigsaw puzzle, Generative Adversarial Network, Hungarian Algorithm, Spatial Reasoning
\end{keywords}
\section{Introduction} 
Historically, jigsaw puzzles were introduced by the British cartographer John Spilsbury in 1760 as a children’s game to develop cognitive reasoning. Since then, the computational problem of solving a puzzle has found several applications 
such as image reconstruction~\cite{legopaper}, assembling of broken objects~\cite{Palmas:etal,elnaghy2019complementarity}, molecular docking~\cite{teodoro2001molecular}, 
and fresco reconstruction~\cite{BrownSIGGRAPH2008}. Recently, the visual problem of solving a puzzle from its unordered pieces has attracted a revamped interest in the Machine Learning and Computer Vision communities~\cite{choCVPR10probjigsaw,SholomonCVPR2013,SantaCruzCVPR17} 
as an example of a spatial reasoning task that is comparatively easy for humans. 


Unfortunately, despite its interest and importance, the Jigsaw puzzle problem is computationally intractable~\cite{DemaineGAC07}. Hence, image reassembly methods require the development of heuristics to efficiently place the pieces by taking into account not only visual information such as texture, color, but also the geometrical information such as the shape and orientation of each piece. Such features are used to compute pairwise affinities between pieces \cite{SholomonCVPR2013,SonECCV2014} for a greedy piece placer optimization \cite{pomeranzCVPR11greedy}. 
Alternatively, the current application of neural networks have several drawbacks which limit their applicability including ability to handle different sized puzzles and lack of global information \cite{bridger2020solving, NorooziECCV2016, RafiqueICACS19ganplacement}.

\begin{figure*}[t]
    \centering
    \includegraphics[width=\linewidth] {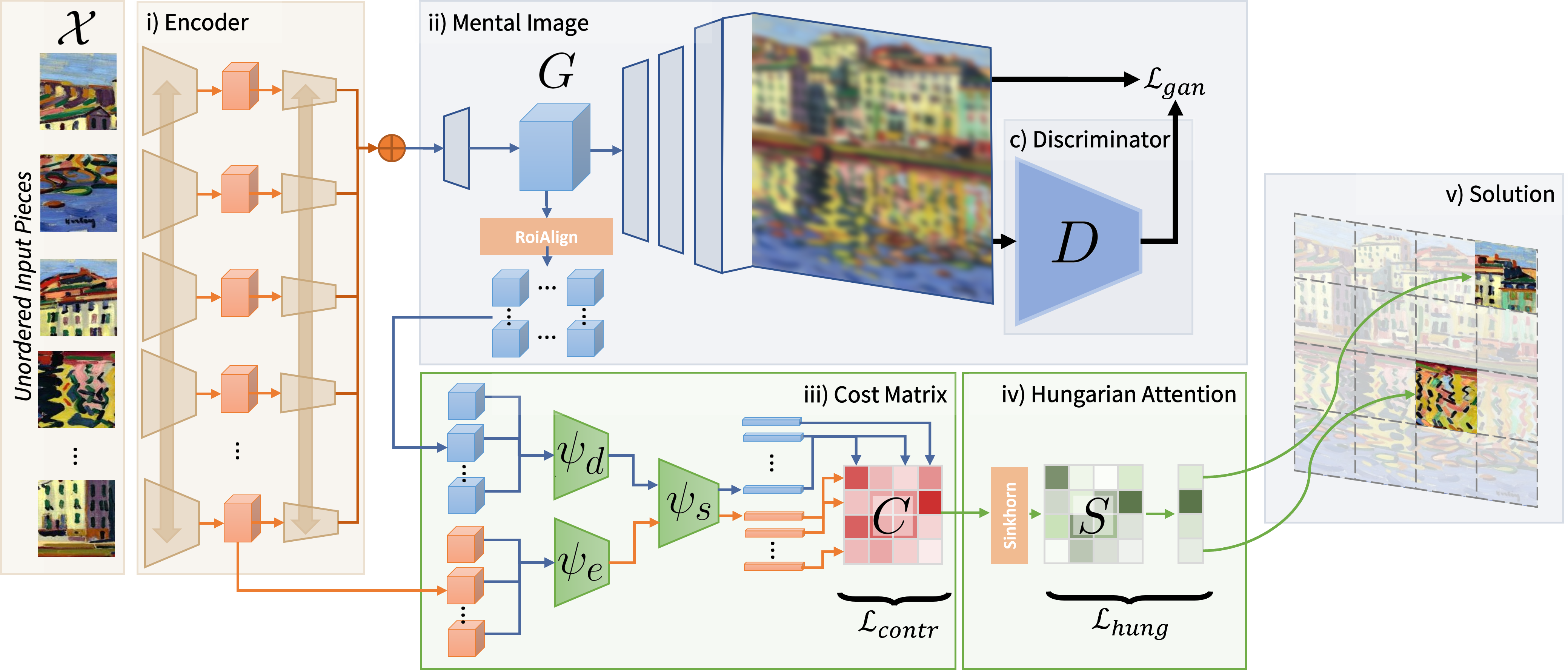}
    \caption{\NETNAME takes as input a set of unordered pieces ($\mathcal{X}$), which is passed through an encoder (i) and then pooled to produce a latent vector. The latent vector is used to generate the \textit{mental image} (ii) using the generator ($G$) and fine-tuned using a discriminator ($D$). An intermediate encoding layer of the generator is cropped using RoiAlign to act as targets. (iii) Both the cropped targets and pieces are encoded through embedding networks ($\psi_d$ and $\psi_e$) and then with a common encoder $\psi_s$. To produce a cost matrix ($C$) dot product is used.  Hungarian attention (iv) is used to solve for the final location where $C$ is normalized through the iterative Sinkhorn normalization ($S$) to obtain a doubly stochastic matrix and in turn solve the assignment problem producing the permutation of the solution (v).}
    \label{fig:model}
\end{figure*}

In contrast, we aim to find a global solution by first generating the whole image from the pieces and then estimating their positions. Specifically, we propose GANzzle including a Generative Adversarial Network (GAN) based on a multi-encoder single-decoder architecture, which can then place pieces by evaluating the matching of segments and target slots in the generated image, using a differentiable hungarian attention (Fig.~\ref{fig:model}). 
The contributions are therefore three-fold: 1) A many-to-one GAN Architecture for recovering a global image from its  pieces; 2) Dynamic puzzle size solver using Hungarian attention and contrastive loss; 3) Two new large-scale puzzle solving datasets, named \textit{PuzzleCelebA} and \textit{PuzzleWikiArts}.


\section{Related Work}
Two main computational approaches have been explored to solve the image jigsaw problem optimization and deep learning methods. 

\noindent\textbf{Optimization:} exploit edges of pieces to formulate a compatibility metric which can then be optimized. Cho et al. ~\cite{choCVPR10probjigsaw} formulate the puzzle-solving problem through a graphical model for label assignment to slots, then the belief propagation optimization shares neighbor information to already placed pieces. By considering the continuity of gradients in adjacent pieces, \cite{GallagherCVPR20112} casts puzzles as a minimum spanning tree problem where edges represent spatial relationships between pieces. To overcome seed sensitivity of the aforementioned methods, \cite{pomeranzCVPR11greedy} iteratively refines the seed to be the largest found segment. Placement and segmentation methods are based on the best-buddy heuristic where pieces agree on being neighbors. Jointly with a relative placement of pieces, \cite{PaikinCVPR2015} proposes that the first placed piece should be distinctive and in a distinctive area. In contrast, we propose a global solution that can support local piece placements and therefore be performed efficiently in a single forward pass of a neural network without iterative pair-wise refinement.

\noindent\textbf{Deep Learning:} Similar to optimization methods Zhang et al. \cite{zhang2018ICLR} considered a learnable cost function. An alternating optimization infers the correct permutation and a suitable cost matrix assessing pairwise relationships. The generative model in \cite{RafiqueICACS19ganplacement} outputs a placement vector, i.e. a vector whose elements indicate where pieces should be placed, while the discriminator outputs the probability of that vector representing a real-placement. In contrast, we pose a synthesized image is a simpler solution for solving for the piece assignment. \cite{bridger2020solving} tries to infill using a GAN between pairs of pieces and therefore solving the assembly problem. While this may seem similar to \NETNAME, its solution is pair-wise and takes ${\sim}60s$ to solve a puzzle of 70 pieces, considerably slower than the proposed approach. The regression of the applied permutation matrix, which is intrinsically discrete and hence a non-differentiable problem, is relaxed in \cite{SantaCruzCVPR17}. At inference time, the closest permutation matrix is selected via linear integer programming. In Deepzzle \cite{paumard2020deepzzle}, a convolutional network predicts relative placement of pieces with respect to a central anchor. By assuming the independent placement of consecutive pieces, the algorithm builds an assembly tree where edges encode log probabilities of placement. \cite{li2021jigsawgan} employs a generative adversarial network to include semantic information in the placement of pieces, where a classification branch predicts which permutation out of a fixed set has been applied to produce the associated flow-based warp on features. While in \cite{li2021jigsawgan} the adversarial branch aids the classification of the given permutation by projecting it in the image space, in \NETNAME, the generator is learning to permute pieces correctly. As a benefit, our solution can cope with an arbitrary permutation of pieces.

\section{Piece assignment with global information}
The \NETNAME  model takes as input a set of $n$ pieces (or image patches) $\mathcal{X} = \{X_0, X_1, \ldots, X_n \}$, $X_i \in {\rm I\!R}^{pw \times ph}, i=1, \ldots, n$ and aims to infer piece locations supported by the reconstructed target image $\mathcal{Y} \in {\rm I\!R}^{tw \times th}$, with $p$ and $t$ being the piece and target dimensions respectively.
At first, for each input piece $X_i$, we learn an embedding representation, then a pooling strategy  outputs a fixed-sized encoding which is fed to the decoder to generate a synthetic target image (fig.~\ref{fig:model}i \& fig.~\ref{fig:model}ii, sec.~\ref{sec:mesd}).
To solve for piece positions, we learn to match the pieces to targets within the global encoding. In the feature space, a cost matrix between patches and target slots accounts for the cost of their assignment  (fig.~\ref{fig:model}iii). Therefore, we optimize the assignments using the differentiable Hungarian attention algorithm which makes the model attend only relevant assignment information (fig.~\ref{fig:model}iv, sec. \ref{sec:match}).


\subsection{Learning global side information}\label{sec:mesd}
To create the mental image, i.e., a rough estimate of the global solution, we use an encoder-decoder architecture to generate an image from the pieces. We modify traditional network to accept $n$ pieces as inputs, similar to multi-view approaches \cite{suICCV15mvcnn}, where each piece is passed through the multi-encoder which uses shared weights across the pieces.
Contrary to the fixed-order assumption of \cite{suICCV15mvcnn}, the unordered nature of jigsaw puzzles requires learning piece embeddings and gather them through average-pooling to generate a single encoding vector across all the pieces. 
The pooled encoding is then passed to the decoder. As with prior work, we found the discriminator's feedback increased high-frequency information, creating more representative synthetic images necessary to guide the subsequent matching of pieces. We opt for MSG-GAN style of approach to provide supervision to the GAN at multiple levels. The generator $G(\cdot)$ and the discriminator $D(\cdot)$ are trained in the standard min max fashion:


\begin{equation}\label{eq:ganloss}
    \mathcal{L}_{gan}(G, D)= {\rm I\!E}_{y}[\text{log}\;D(y)] + {\rm I\!E}_{x}[\text{log}\;(1-D(G(x)))],
\end{equation}
where the first and the second expectation are computed on target images $\mathcal{Y}$ and images generated from pieces $\mathcal{X}$, respectively. While the discriminator maximizes eq. (\ref{eq:ganloss}), the generator minimizes it.
As in MSG-GAN, the equations are extended over all resolutions to provide multi-scale gradients:
\begin{equation*}
    \mathcal{L}_{GAN} = \min_{G}\;\max_{D} \sum^{L}_{l=1} L_{gan}(G^l,D^l) + \lambda_{p} \mathcal{L}_{mse}(G^l),
\end{equation*}
where $G^l$ is the RGB-converted intermediate representation of the generator at layer $l$ (depth $L$), $D^l$ the corresponding discriminator. A pixel-wise mean squared error term $\mathcal{L}_{mse}(\cdot)$ is included weighted by $\lambda_p$, balancing the reconstruction term with the refinements introduced by the discriminator. 

Using MSG-GAN removes the expectation of generating larger images than the sum of input dimension pieces. We exploit this to incrementally grow the GAN during training. Furthermore, we group training into batches based on the jigsaw complexity ($2\times2,3\times3,...,n\times n$). Hence, the network does not need to handle a dynamic number of pieces, while remaining puzzle size agnostic at test time.


\subsection{Piece assignment}\label{sec:match}
From the pieces $\mathcal{X}$, we construct a similarity matrix between an intermediate encoding of $X_i$ and the placement target slots of the intermediate embedding of the decoder partitioned by RoiAlign. Both intermediate encodings are passed through shallow networks $\psi_{e}$ and $\psi_{d}$ for piece and target slots respectively, to avoid degradation of the GAN. Then, they are jointly embedded through $\psi_{s}$ to align the embedding spaces.
Hence, the similarity matrix is computed as dot product of all possible piece-slot pairs at runtime, making it dynamic to the size of the puzzle.
A contrastive loss enforces the feature space to have similar embeddings for piece-slot correct pairs while pushing apart non-corresponding pairs:
\begin{equation*}
    \mathcal{L}_{contr}{=}{-}\mathbb{E}_i\left[\log\frac{exp\left(\psi_s^i \cdot \psi_s^j / \tau\right)}{exp(\psi_s^i \cdot \psi_s^j / \tau){+}\sum_{k \neq j} exp(\psi_s^i \cdot \psi_s^k / \tau )}\right]
\end{equation*}
with $\psi_s^i$ and $\psi_s^j$ embeddings of considered piece $i$ and its corresponding slot $j$, $\tau$ the temperature parameter.

Assignments based on the cost matrix could then be efficiently computed by employing the Hungarian Algorithm \cite{kuhn1955hungarian}. However, the algorithm is non-differentiable due to the discrete nature of assignments. 
Therefore, we employ Hungarian attention (HA) \cite{yu2019learning} to learn the assignment task in a supervised way, where a continuously relaxed assignment problem is shaped such that its Hungarian solution is the desired one. Initially, the cost matrix ($C$) is normalized through the iterative Sinkhorn normalization to obtain a doubly stochastic matrix:
\begin{equation}
    S^0(C) = \exp(C)\\
\end{equation}
\begin{equation}
    S^l(C) = F_c\left(F_r\left(S^{l-1}\right)\right)\\
\end{equation}
\begin{equation}
    S(C) = \lim_{l \rightarrow\infty} S^l\left(C\right),
\end{equation}
where $F_c$ and $F_r$ are the row and column-wise normalization $F_c(C) = C \oslash \left( \mathbf{1}_N\mathbf{1}_N^TC \right)$ and $F_r(C) = C \oslash \left( C\mathbf{1}_N\mathbf{1}_N^T \right)$ respectively, with $\oslash$ denoting the  element-wise division and $1_N$ the $n$-dimensional unit column vector.
Therefore, its assignment counterpart is determined via HA matching $\mathrm{Hung}\left(S\right)$. Thus, a hard attention mask is generated by comparing it to the ground-truth assignment $\mathbf{S}^G$ through a element-wise logic-OR operator:
\begin{equation*} \label{hungarianMask}
    \mathbf{Z} = OR\left(\mathrm{Hung}\left(\mathbf{S}\right), \mathbf{S}^G\right).
\end{equation*}
Note that the hard mask catches most relevant elements in the matrix to avoid overconfidence, however, both correct and misplaced pieces are modeled.

The binary cross-entropy loss with respect to the ground-truth assignment matrix is attended through the mask:
\begin{equation*} \label{hungarianLoss}
    \mathcal{L}_{hung} = \sum_{i, j  \in \left[n\right]} \mathbf{Z}_{ij}\left(\mathbf{S}_{ij}^G \log \mathbf{S}_{ij} + \left(1-\mathbf{S}_{ij}^G\right)\log\left(1-\mathbf{S}_{ij}\right)\right),
\end{equation*}
where  $\left[n\right]$ is the set of indexes from $1$ to $n$.

By optimizing the above permutation loss, our model learns to correctly match the Hungarian's assignment computed from $S$ to the ground truth permutation. At inference time, the estimated assignment is hence the Hungarian binarization of the doubly stochastic matrix $\mathrm{Hung}\left(S\right)$.

The complete loss for the \NETNAME model is therefore:
\begin{equation}
    \mathcal{L} = \mathcal{L}_{GAN} + \mathcal{L}_{hung} + \mathcal{L}_{contr}. 
\end{equation}

We additionally consider a model that exploits only HA without the generated mental image as permutation based method referred to Hung-perm (See supp. mat. for full details).


    

\begin{table*}[t]
    \scriptsize
    \centering
    \begin{tabularx}{\linewidth}{X  c@{\hskip 0.3in}  c@{\hskip 0.3in} c@{\hskip 0.3in} c@{\hskip 0.5in} c@{\hskip 0.3in}  c@{\hskip 0.3in}  c@{\hskip 0.3in} c}
    \toprule
    \textbf{Dataset} & \multicolumn{4}{c}{\textbf{PuzzleCelebA}} & \multicolumn{4}{c}{\textbf{PuzzleWikiArts}} \\
    \midrule
    \textbf{} & \textbf{6x6} & \textbf{8x8} & \textbf{10x10} & \textbf{12x12} & \textbf{6x6} & \textbf{8x8} & \textbf{10x10} & \textbf{12x12} \\
    \midrule
    
    Paikin and Tal~\cite{PaikinCVPR2015}& 99.12 & 98.67 & 98.39 & 96.51  & 98.03 & 97.35 & 95.31 & 90.52 \\
    Pomeranz et al.~\cite{pomeranzCVPR11greedy} & 84.59  & 79.43  & 74.80  &  66.43  & 79.23  & 72.64  & 67.70  & 62.13 \\
    Gallagher ~\cite{GallagherCVPR20112} &  98.55  &  97.04  & 95.49  & 93.13  & 88.77  & 82.28  & 77.17  & 73.40  \\
    \midrule
    PO-LA~\cite{zhang2018ICLR}  & 71.96  & 50.12   & 38.05  & -& 12.19  & 5.77  & 3.28  & -\\
    \hdashline
    Hung-perm  & 33.11  & 12.89  & 4.14  & 2.18  & 8.42  & 3.22  & 1.90  & 1.25 \\
    \NETNAME-Single (Ours) & 71.00 & 51.81 & \textbf{43.74} & - & 11.78 & 6.23 & \textbf{8.97} & - \\
    \NETNAME (Ours) & \textbf{72.18} & \textbf{53.26} & 32.84 & \textbf{12.94} & \textbf{13.48} & \textbf{6.93} & 4.10 & \textbf{2.58}  \\
    
    \bottomrule
    \end{tabularx}
    \centering
    \caption{\label{tab:comparison}Results for direct comparison accuracy on PuzzleCelebA and PuzzleWikiArts. We directly compare against deep method (\cite{zhang2018ICLR}) and without mental image (Hung-perm) for similar computational performance and include optimization methods \cite{PaikinCVPR2015,pomeranzCVPR11greedy,GallagherCVPR20112} for complete comparison. We note, that in contrast to \NETNAME, \cite{zhang2018ICLR} is trained one model per size. }
\end{table*}
\begin{table*}[t]
    \scriptsize
    \centering
    \begin{tabularx}{\textwidth}{X c@{\hskip 0.5in}c@{\hskip 0.5in}c c@{\hskip 0.5in}c@{\hskip 0.5in}c c@{\hskip 0.5in}c@{\hskip 0.5in}c}
    \toprule
    \multirow{2}{*}{\textbf{Model}} &
    \multicolumn{3}{c}{\textbf{Missing (\%)}} & \multicolumn{3}{c}{\textbf{Noisy ($\mathbf{\sigma}$)}} & \multicolumn{3}{c}{\textbf{Eroded (px)}}\\
    \cmidrule(lr){2-4}
    \cmidrule(lr){5-7}
    \cmidrule(lr){8-10}
    & \textbf{10\%} & \textbf{20\%} & \textbf{30\%} & \textbf{0.05} & \textbf{0.1} & \textbf{0.2} & \textbf{1} & \textbf{2} & \textbf{5}\\
    \midrule
    
    Paikin and Tal~\cite{PaikinCVPR2015} & - & - & - & 51.51  & 7.73  & 3.31  & 2.82  & 2.77  & 2.79 \\
    Pomeranz et al.~\cite{pomeranzCVPR11greedy} & 52.43  & 24.26  & 25.99  & 87.84  & 89.63  & 91.50  & 6.01  & 16.30  & 15.15  \\
    Gallagher ~\cite{GallagherCVPR20112} & 79.68  & 66.02  & 51.17  & 96.39  & 98.34  & 97.75 & 32.55  & 18.59  & 6.27 \\
    \midrule

    PO-LA~\cite{zhang2018ICLR} & \textbf{64.35}  & \textbf{60.10}  & \textbf{58.60}  & \textbf{69.87}  & \textbf{65.30}  & \textbf{49.85}  & 23.81  & 10.93  & 4.84 \\
    \hdashline
    Hung-perm & 29.79  & 26.45  & 23.88  & 31.84  & 29.01  & 21.45  & 25.45  & 26.01  & \textbf{9.50}  \\
    \NETNAME-Single & 59.00 & 46.24 & 37.54 & 63.81 & 48.95 & 10.13 & \textbf{28.75} & \textbf{39.21} & 9.23\\
    \NETNAME & 58.50 & 44.70 & 35.01 & 64.51 & 37.72 & 6.81 & 28.59 & 35.47 & 4.70\\
    
    \bottomrule
    \end{tabularx}
    \centering
    \caption{\label{tab:noisy-pieces}Comparison of missing pieces (except \cite{PaikinCVPR2015}), Gaussian noise and eroded pieces on a $6{\times}6$ puzzle for PuzzleCelebA.
    }
\end{table*}

\section{Evaluation}
We evaluate on the PuzzleCelebA ($30k$ images) and PuzzleWikiArts ($63k$ images) datasets derived from their respective CelebA~\cite{liuICCV15celebA} and part of WikiArts~\cite{TanTIP19artgan}. We provide train and testing split ($80{-}20\%$) and the permutation for puzzle sizes $[2,4,6,8,10,12]$. 
For evaluation, we use standard \textbf{direct comparison metric}~\cite{choCVPR10probjigsaw} where an assignment is correct if it is placed in the correct position. Full details of the datasets and extended evaluation (inc. Neighbor comparison metric) can be found in the supplementary material.\footnote{Supp. Mat. is available at \url{https://github.com/IIT-PAVIS/GANzzle}}

We show Direct Accuracy for both datasets in Table~\ref{tab:comparison}, for optimization methods \cite{PaikinCVPR2015,pomeranzCVPR11greedy,GallagherCVPR20112}, deep method \cite{zhang2018ICLR} and based only on patch embedding and Hungarian attention (Hung-perm). For PO-LA, as it is fixed size we train one model per size, however, $12{\times}12$ was limited due to memory, we include the single size {\NETNAME}(-Single) for direct comparison. 
Our method generalizes across sizes with similar performance to the single version up to $10{\times}10$. In comparison,  \NETNAME is competitive with \cite{zhang2018ICLR} and out performs Hung-perm on large sizes benefiting from the mental image. 

\begin{figure}[t!]
    \centering
    \includegraphics[width=0.85\linewidth] {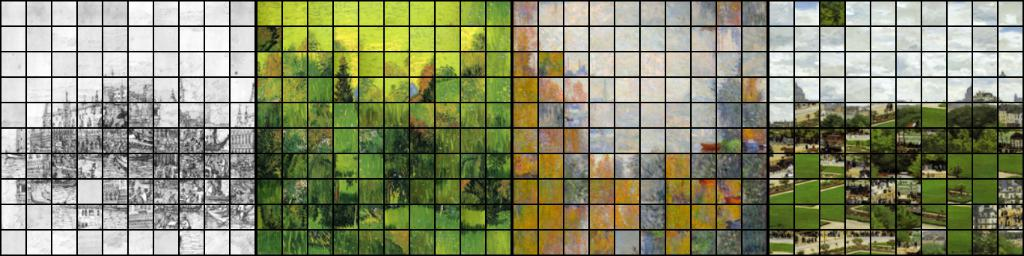}
    \includegraphics[width=0.85\linewidth] {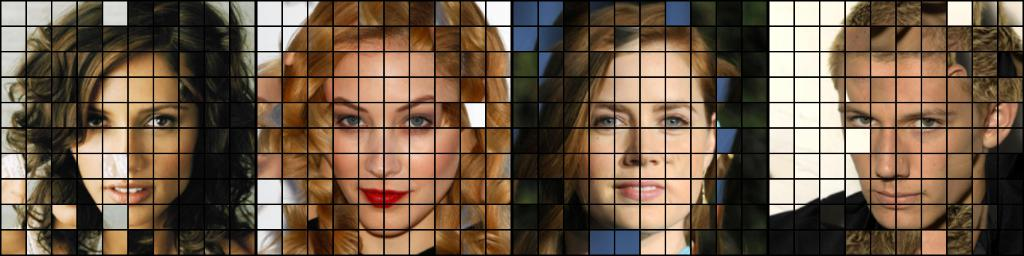}
    \caption{Qualitative results of \NETNAME for $10 \times 10$ on PuzzleWikiArts and PuzzleCelebA (see Supp.).
    }
    \label{fig:results}
\end{figure}

In addition, we evaluate on PuzzleCelebA with missing, noisy, and eroded (missing border) pieces for $6{\times}6$ puzzles in Table~\ref{tab:noisy-pieces}. For missing pieces we are competitive against \cite{zhang2018ICLR} and \cite{pomeranzCVPR11greedy}.  However, \NETNAME struggles in contrast to other methods with additive noise, although it outperforms Hung-Perm benefiting from the mental image. While for eroded pieces \NETNAME outperforms \cite{zhang2018ICLR} and Hung-Perm, in addition for small erosion ($1$\&$2$px) are competitive with optimization methods. \NETNAME consistently performs similar to the single size version highlighting the generalization. Limitations emerge with challenging pieces, i.e., pieces with similar content, as they can be swapped. The qualitative analysis in Fig.~\ref{fig:results} shows how \NETNAME is able to resolve for the structure.


\section{Conclusion}
Our global to local solution \NETNAME is able to learn to position jigsaw puzzle pieces correctly with a single forward pass of the network. 
The use of Hungarian attention is dynamic to jigsaw puzzle sizes and successfully solves for the location of pieces using a singular model.  We have shown results on two new datasets to standardize the jigsaw puzzle solving problem, where we perform competitively with deep methods but overcome the single-size model problem.

\vfill\pagebreak

\bibliographystyle{IEEEbib}
\bibliography{puzzle}
\appendix
\section*{Appendix}

In this appendix, we present extended details on the dataset and evaluation. 
In section~\ref{sec:datasets} we provide quantitative statistics of the two proposed datasets PuzzleCelebA and PuzzleWikiArts. In section~\ref{sec:baselines}, we define the implementation details of baselines and an analysis of their performance through the two evaluation metrics, Direct Accuracy (Tables ~\ref{tab:celebA-direct}, ~\ref{tab:wiki-direct}) and Neighbor accuracy (Tables~\ref{tab:celebA-neigh},~\ref{tab:wiki-neigh}) as well as the performance on missing, noisy or eroded pieces (Table ~\ref{tab:noisy-pieces}). We additionally include analysis on puzzle sizes $2$ and $4$ not shown in the primary text.
We also provide additional qualitative analysis over multiple puzzles for both the PuzzleCelebA and PuzzleWikiArts datasets in Figure~\ref{fig:qual_celeb} and Figure~\ref{fig:qual_wiki}.

\begin{figure*}[t]
     \centering
     \begin{subfigure}[b]{0.49\linewidth}
         \centering
         \includegraphics[width=\linewidth]{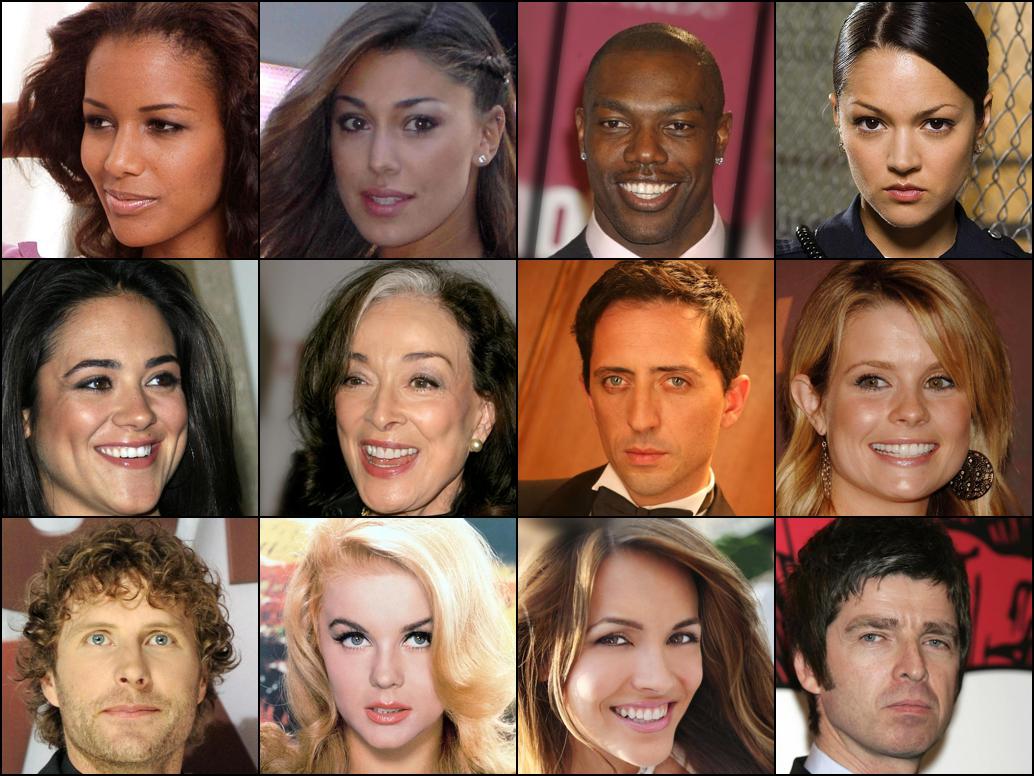}
         \caption{ }
         \label{fig:dataset_celeba}
     \end{subfigure}
     \hfill
     \begin{subfigure}[b]{0.49\linewidth}
         \centering
         \includegraphics[width=\linewidth]{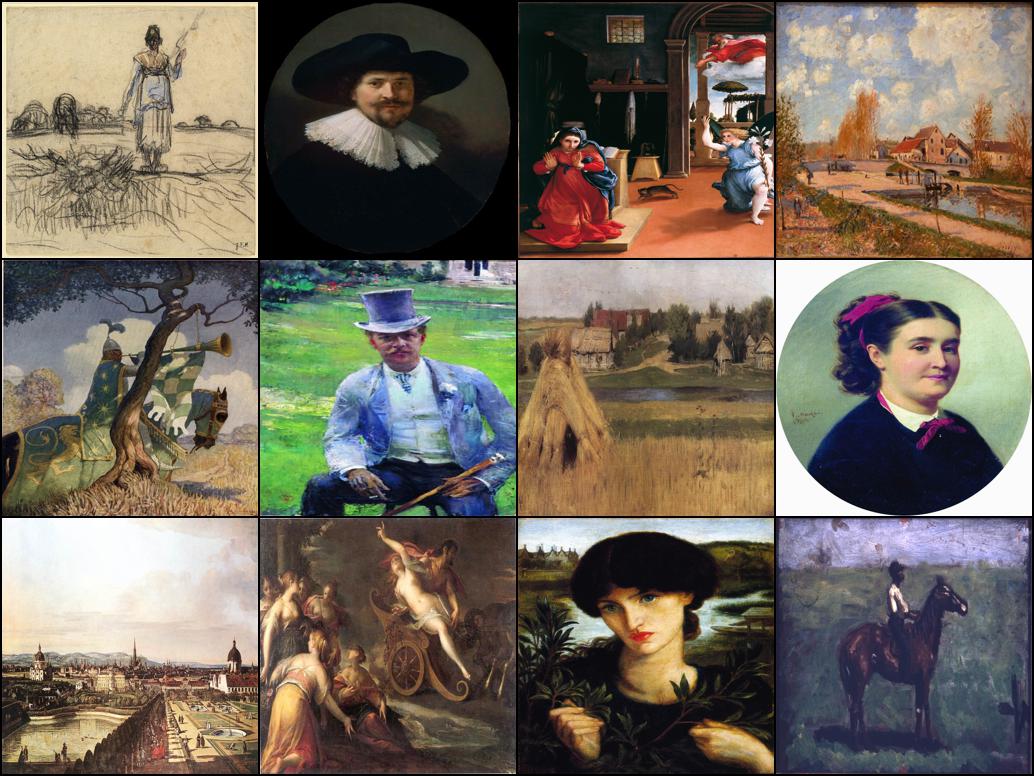}
         \caption{ }
         \label{fig:dataset_wikiart}
     \end{subfigure}
        \caption{Example puzzles from (a) PuzzleCelebA and (b) PuzzleWikiArts}
        \label{fig:dataset}
\end{figure*}

\section{Puzzle Datasets}\label{sec:datasets}
We propose two datasets providing different challenges. Firstly, a visually simple (and consistent) dataset based on CelebA~\cite{liuICCV15celebA}, which has been demonstrated that generative methods are able to synthesize with high accuracy. Secondly, an arts-centered dataset based on WikiArts~\cite{TanTIP19artgan}, this provides a challenging environment for generalization of methods across different styles and content.
In contrast to prior approaches both of these datasets are large (thousands of images), therefore providing a testing ground for both puzzle and spatial reasoning problems.\footnote{Datasets available at: \url{https://github.com/IIT-PAVIS/GANzzle}}
For each dataset we provide a text file split for training and test. To enable fair comparison between different approaches, for test samples we provide fixed shuffling permutations.

\subsection{PuzzleCelebA}
To evaluate our model, we relied on a high quality version of CelebA  \cite{CelebAMask-HQ}, a large-scale face image dataset (see Fig.~\ref{fig:dataset_celeba}). Counter intuitively, faces represent a challenging testbed for jigsaw puzzle solving. Two sources of ambiguities are concurrently involved: faces are highly symmetrical and profile pictures are characterized by blurred (or plain) background. We noticed the consistent structure of images to ease the generation of guiding images to match against. From the $30k$ images we applied a random $80-20\%$ train-test split and for each image we generate six grid sizes, i.e., complexities, $[2,4,6,8,10,12]$ where each generated puzzle is randomly shuffled. We provide the $35,994$ test puzzle permutations for comparative evaluation.

\subsection{PuzzleWikiArts}
We build from a subset of the WikiArts dataset \cite{TanTIP19artgan}, proposing a split and different grid sizes for evaluation. We select art style categories that avoid highly complex symmetries that make placing pieces ambiguous, therefore the dataset contains varying difficult examples including more unique humanoid structures as well as patterns that will challenge puzzle solving algorithms with near duplicate pieces (see Fig.~\ref{fig:dataset_wikiart}). We take the $63,126$ images of the dataset and split them randomly into an $80-20\%$ train-test split resulting in $50,502$ training images. Similarly to CelebA jigsaw data, for each image we generated three grid sizes. A total of $37,875$ test puzzles is present.

\section{Baseline details}\label{sec:baselines}
We compared our proposed with representative methods of different approaches. To this end, we considered three different optimization-based algorithms: \textbf{Gallagher} \cite{GallagherCVPR20112}, \textbf{Pomeranz and Tal} \cite{pomeranzCVPR11greedy} and \textbf{Paikin et al.} \cite{PaikinCVPR2015}. For deep learning solutions, we compare to Permutation-Optimization with Linear Assignment (\textbf{PO-LA}) \cite{zhang2018ICLR} from set representation. Furthermore, we considered a deep learning baseline based on Hungarian attention \cite{yu2019learning}, here denoted as \textbf{hung-perm}.

We built on the original implementations of chosen optimization approaches: Paikin and Tal \cite{PaikinCVPR2015}, Pomeranz and Tal  \cite{pomeranzCVPR11greedy} and Gallagher \cite{GallagherCVPR20112}. Hyper-parameters are set to those of the original proposals. Test jigsaw with square 32 pixels patches are fed to solving algorithms, Predicted reordering permutations are hence retrieved. Due to limited and controlled evaluation protocols of classic methods, we found optimization approaches to fail with black tiles that could randomly be present in images. Indeed, original implementations do not characterize missing pieces. Therefore, a major adaptation was needed to set apart black pieces from missing ones. To this end, we set the color of the patch middle point to white. Unfortunately, we could not evaluate such approaches for low-size puzzles, due to instability of implementations.

\begin{table}[b!]
    \centering
    \begin{tabularx}{\linewidth}{X c}
    \toprule
    \textbf{Model} & \textbf{Average Sample (ms)} \\
    \midrule
    Paikin and Tal \cite{PaikinCVPR2015}   & 27.47 $\pm$ 7.70  \\ 
    Pomeranz et al. \cite{pomeranzCVPR11greedy}  & 221.64 $\pm$ 300.79 \\
    Gallagher \cite{GallagherCVPR20112} & 235.19 $\pm$ 358.72 \\
    \midrule
    PO-LA \cite{zhang2018ICLR}& 22.38 $\pm$ 8.08 \\
    \hdashline
    Hung-Perm & 9.97 $\pm$ 1.38  \\ 
    Ganzzle & 25.16 $\pm$ 1.1  \\
    \bottomrule
    \end{tabularx}
    \centering
    \caption{\label{tab:compuational-time} Computation time requirements in ms for the different approaches on a $6 \times 6$ puzzle.}
\end{table}

We adapted PO-LA \cite{zhang2018ICLR} algorithm to handle the large cardinality of puzzles pieces. Specifically, owing to the algorithm's memory requirements, we rescaled tiles to $16\times16$ and set the hidden size of the network producing the ordering cost to $64$. Adam \cite{kingma2014Adam} with a fixed learning rate of $1e-3$ is employed for 20 epochs using a batch size of 32. The PO-LA approach was not assessed for larger sizes due to the computational limitations of its memory footprint.

In hung-perm, a ResNet-50 backbone encodes tiles independently. For each tile, the 2048-dimensional embedding is hence mapped to a a vector of size 256 with a fully connected linear layer. Finally, after gathering of all embeddings by concatenation, the model outputs the 1-1 assignment problem to optimize for ,i.e., the assignment cost matrix.
Similary to \NETNAME, the predicted permuation is obtained as hungarization of the associated doubly stochastic matrix. The model is trained end-to-end via Hungarian attention for 150 epochs. Adam with learning rate of 0.01 and batch size 64 is employed.
In deep learning solutions, we trained a model for each size. 

\section{Implementation Details}
For both baseline and GANzzle experiments we conduct training on $20$ core Intel Xeon, 394Gb RAM and 4x NVIDIA Tesla V100 16Gb. We pre-train the GANzzle GAN for 144hrs (72hrs with out discriminator and 72hrs with), we note that WikiCelebA converges significantly faster and therefore simpler datsets than PuzzleWikiArts could be trained with a much shorter time. We then train the full GANzzle model on WikiCelebA for 24hrs, and  72hrs on PuzzleWikiArts. We use the Adam optimizer with a learning rate of $1e-3$ for the Generator ($G$) and $4e-3$ for the Discriminator ($D$), in joint training we use $1e-3$ for the learning rate.

\section{Computational Time}
We compare the computation wall time averaged over $24$ samples in Table~\ref{tab:compuational-time}. It can be seen against \cite{pomeranzCVPR11greedy} and \cite{GallagherCVPR20112} that the deep learning methods (including \NETNAME) have a significant computational time gain. In contrast, Paikin and Tal has comparable time requirements, however, it should be noted that as the puzzle size increases the time increases too. Deep learning methods are largely similar, with the minimal (without GAN) Hung-Perm taking half the computational time.

\begin{table*}[h!]
\begin{minipage}{1.0\textwidth}
    \scriptsize
    \centering
    \begin{tabularx}{\textwidth}{X c c c c c c}
    \toprule
    \textbf{Model} & \textbf{2x2} & \textbf{4x4} & \textbf{6x6} & \textbf{8x8} & \textbf{10x10} & \textbf{12x12} \\
    \midrule
    Paikin and Tal \cite{PaikinCVPR2015} & - & - & 98.03  & 97.35  & 95.31  & 90.52 \\
    Pomeranz et al. \cite{pomeranzCVPR11greedy} & - & 83.34  & 79.23  & 72.64  & 67.70  & 62.13  \\
    Gallagher \cite{GallagherCVPR20112} & 98.10  & 93.94 & 89.47  & 83.22  & 78.25  & 73.40 \\
    Hung-perm & 47.09  & 30.37  & 8.42  & 3.22  & 1.90  & 1.25  \\
    PO-LA \cite{zhang2018ICLR} & 85.67  & 34.58   & 12.19  & 5.77  & 3.28  & -\\
    \midrule
    \NETNAME-Single (Ours)& - & 25.34 & 11.78 & 6.23 & 8.97 & - \\
    \NETNAME (Ours) & 76.86 & 29.72 & 13.48 & 6.93 & 4.10 & 2.58 \\
    \bottomrule
    \end{tabularx}
    \centering
    \caption{\label{tab:wiki-direct}Direct accuracy comparison on PuzzleWikiArts.}
\end{minipage}
\hfill

\begin{minipage}{1.0\textwidth}
    \scriptsize
    \centering
    \begin{tabularx}{\textwidth}{X c c c c c c}
    \toprule
    \textbf{Model} & \textbf{2x2} &  \textbf{4x4}  & \textbf{6x6} & \textbf{8x8} & \textbf{10x10} & \textbf{12x12} \\
    \midrule
    Paikin and Tal \cite{PaikinCVPR2015} & - & - & 99.37  & 99.09  & 98.23  & 95.97  \\
    Pomeranz et al. \cite{pomeranzCVPR11greedy} & - & 94.83  & 93.39  & 89.96  & 87.25  & 84.07  \\
    Hung-perm & 45.36  & 18.72  & 4.25   & 1.97   & 1.43  & 0.90  \\
    PO-LA \cite{zhang2018ICLR} & 82.32   & 26.40  & 7.94  & 4.01   & 2.58  & - \\
    \midrule
    \NETNAME-Single (Ours) & - & 17.96 & 9.53 & 6.30 & 8.21 & - \\
    \NETNAME (Ours) & 72.33 & 22.61 & 11.08 & 7.10 & 5.32 & 4.18 \\
    \bottomrule
    \end{tabularx}
    \centering
    \caption{\label{tab:wiki-neigh}Neighbor accuracy comparison on PuzzleWikiArts.}
\end{minipage}
\hfill

\begin{minipage}{1.0\textwidth}
    \scriptsize
    \centering
    \begin{tabularx}{\textwidth}{X c c c c c c}
    \toprule
    \textbf{Model} & \textbf{2x2} & \textbf{4x4} & \textbf{6x6} & \textbf{8x8} & \textbf{10x10} & \textbf{12x12} \\
    \midrule
    Paikin and Tal \cite{PaikinCVPR2015} & - & - & 99.12  & 98.67  & 98.39  & 96.51  \\
    Pomeranz et al. \cite{pomeranzCVPR11greedy} & - & 84.67  & 84.59  & 79.43  & 74.80  &  66.43  \\
    Gallagher \cite{GallagherCVPR20112} & 99.53  & 95.76  &  90.08  &  77.69  & 66.68  & 53.53  \\
    Hung-perm & 99.95  & 88.05  & 33.11  & 12.89  & 4.14  & 2.18  \\
    PO-LA \cite{zhang2018ICLR} & 99.80  & 92.20  & 71.96   & 50.12   & 38.05  & -\\
    \midrule
    \NETNAME-Single (Ours) & - & 91.05 & 71.00 & 51.81 & 43.74 & - \\
    \NETNAME (Ours) & 99.67 & 87.80 & 72.18 & 53.26 & 32.84 & 12.94 \\
    \bottomrule
    \end{tabularx}
    \centering
    \caption{\label{tab:celebA-direct}Direct accuracy comparison  on PuzzleCelebA.}
\end{minipage}
\hfill

\begin{minipage}{1.0\textwidth}
    \scriptsize
    \centering
    \begin{tabularx}{\textwidth}{X c c c c c c}
    \toprule
    \textbf{Model} & \textbf{2x2} &  \textbf{4x4}  & \textbf{6x6} & \textbf{8x8} & \textbf{10x10} & \textbf{12x12} \\
    \midrule
    Paikin and Tal \cite{PaikinCVPR2015} & - & - & 99.70  & 99.38  & 99.15  & 96.51 \\
    Pomeranz et al. \cite{pomeranzCVPR11greedy} & - & 96.39  & 96.31  & 93.87  & 91.38  & 87.79 \\
    Hung-perm &  99.92  & 84.57 & 22.35 & 7.49 & 2.33 & 0.95 \\
    PO-LA \cite{zhang2018ICLR} & 99.71  & 89.82  & 66.43  & 44.02  & 32.72  & -\\
    \midrule
    \NETNAME-Single (Ours) & - & 88.13 & 64.70 & 44.15 & 37.51 & - \\
    \NETNAME (Ours) & 99.52 & 83.87 & 66.04 & 46.20 & 26.46 & 9.93\\
    \bottomrule
    \end{tabularx}
    \centering
    \caption{\label{tab:celebA-neigh}Neighbor accuracy comparison on PuzzleCelebA.}
\end{minipage}
\hfill

\begin{minipage}{1.0\textwidth}
    \scriptsize
    \centering
    \begin{tabularx}{\textwidth}{X ccc ccc ccc}
    \toprule
    \multirow{2}{*}{\textbf{Model}} &
    \multicolumn{3}{c}{\textbf{Missing (\%)}} & \multicolumn{3}{c}{\textbf{Noisy ($\mathbf{\sigma}$)}} & \multicolumn{3}{c}{\textbf{Eroded (px)}}\\
    \cmidrule(lr){2-4}
    \cmidrule(lr){5-7}
    \cmidrule(lr){8-10}
    & \textbf{10\%} & \textbf{20\%} & \textbf{30\%} & \textbf{0.05} & \textbf{0.1} & \textbf{0.2} & \textbf{1} & \textbf{2} & \textbf{5}\\
    \midrule
    
    Paikin and Tal \cite{PaikinCVPR2015} & - & - & - & 47.58  & 9.83  & 3.36  & 2.81  & 2.78  & 2.79  \\
    Pomeranz et al. \cite{pomeranzCVPR11greedy} & 43.12  & 21.44  & 17.66  & 84.27  & 87.62  & 90.49  & 3.18  & 3.18  & 5.64  \\
    Gallagher \cite{GallagherCVPR20112} & 80.94  & 66.76  & 52.20  & 95.91  & 96.28  & 96.30  & 31.62  & 23.48  & 7.58  \\
    Hung-perm & 7.85   & 7.12  & 6.55  & 6.63  & 5.35  & 4.24  & 3.69  & 4.03  & 3.17  \\
    PO-LA \cite{zhang2018ICLR} & 11.42  & 13.45  & 16.64  & 11.77  & 10.75  & 8.43  & 4.96  & 3.23  & 2.58  \\
    \midrule
    \NETNAME-Single (Ours) & 10.20 & 8.34 & 7.21 & 10.79 & 9.63 & 7.90 & 4.73 & 4.52 & 3.12\\
    \NETNAME (Ours) & 11.60 & 9.44 & 7.88 & 11.80 & 10.22 & 8.19 & 5.07 & 4.99 & 3.29\\
    \bottomrule
    \end{tabularx}
    \centering
    \caption{\label{tab:noisy-pieces}PuzzleWikiArts dataset. Comparison of different noise techniques for pieces using direct accuracy comparison for missing pieces,  additive Gaussian noise and eroded pieces on a $6\times6$ puzzle.}
\end{minipage}
\\
\\
\\
\\
\\
\\
\\
\\
\\
\end{table*}


\begin{table*}
\begin{minipage}{1.0\textwidth}
    \scriptsize
    \centering
    \begin{tabularx}{\textwidth}{X c c c c c c}
    \toprule
    \textbf{Model} & \textbf{2x2} &  \textbf{4x4}  & \textbf{6x6} & \textbf{8x8} & \textbf{10x10} & \textbf{12x12} \\
    \midrule

    Paikin and Tal \cite{PaikinCVPR2015} & - & - & \includegraphics[width=0.110\textwidth]{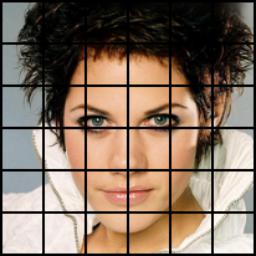}  & \includegraphics[width=0.110\textwidth]{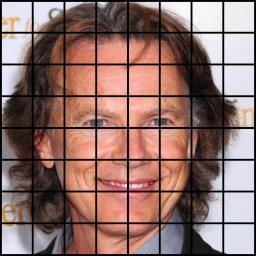}  & \includegraphics[width=0.110\textwidth]{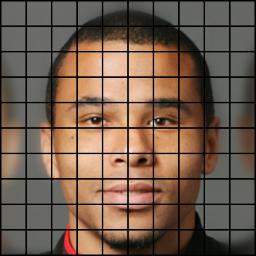} & \includegraphics[width=0.110\textwidth]{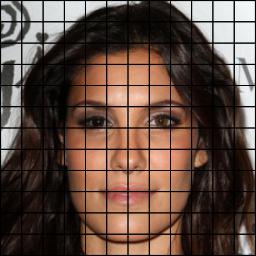} \\
    Pomeranz et al. \cite{pomeranzCVPR11greedy} & - & \includegraphics[width=0.110\textwidth]{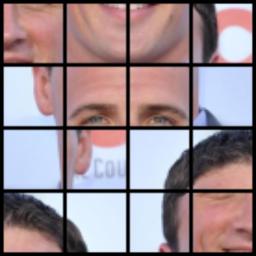} & \includegraphics[width=0.110\textwidth]{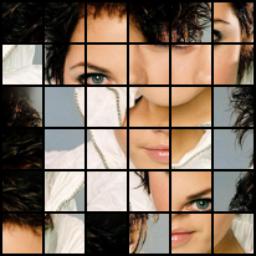}  & \includegraphics[width=0.110\textwidth]{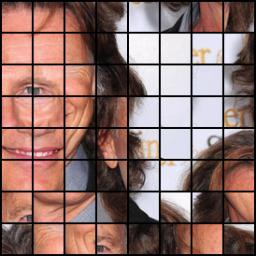}  & \includegraphics[width=0.110\textwidth]{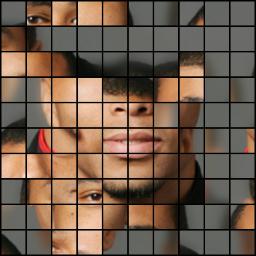} & \includegraphics[width=0.110\textwidth]{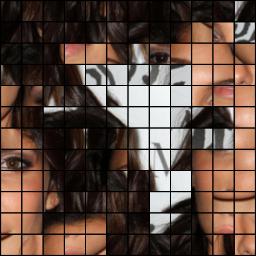}
    \\
    PO-LA \cite{zhang2018ICLR} & \includegraphics[width=0.110\textwidth]{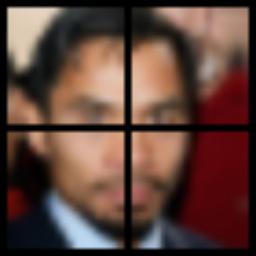} & \includegraphics[width=0.110\textwidth]{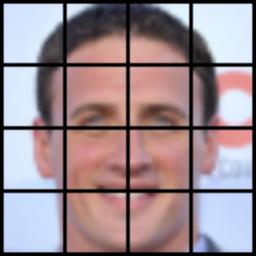} & \includegraphics[width=0.110\textwidth]{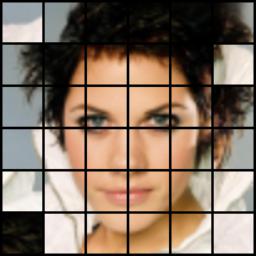}  & \includegraphics[width=0.110\textwidth]{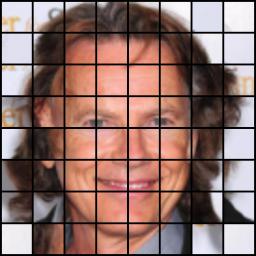}  & \includegraphics[width=0.110\textwidth]{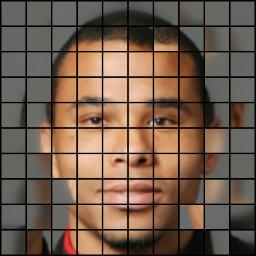} & - \\
    Hung-perm & \includegraphics[width=0.110\textwidth]{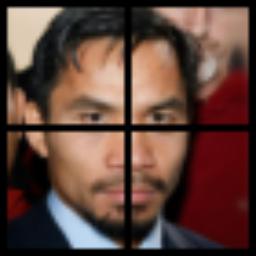} & \includegraphics[width=0.110\textwidth]{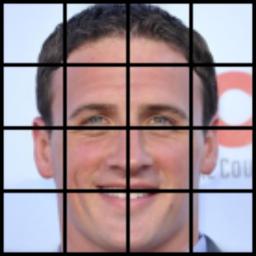}  & \includegraphics[width=0.110\textwidth]{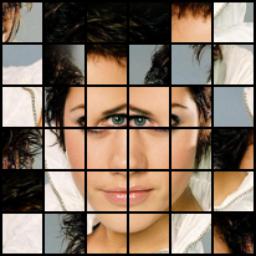}  & \includegraphics[width=0.110\textwidth]{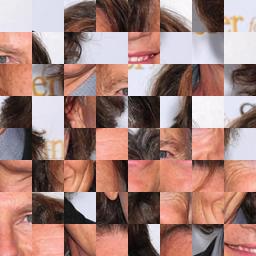} & \includegraphics[width=0.110\textwidth]{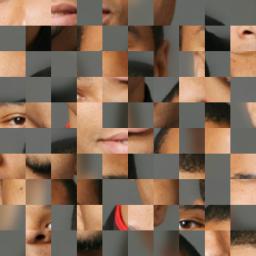} &
        \includegraphics[width=0.110\textwidth]{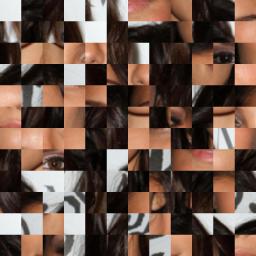} \\
    \midrule
    Ganzzle & \includegraphics[width=0.110\textwidth]{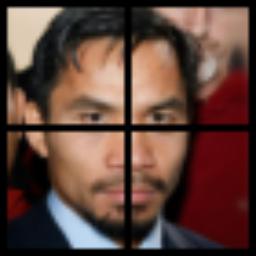} & \includegraphics[width=0.110\textwidth]{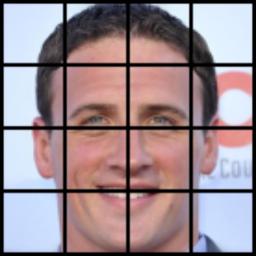} & \includegraphics[width=0.110\textwidth]{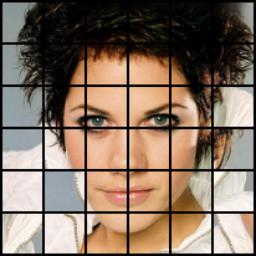}  & \includegraphics[width=0.110\textwidth]{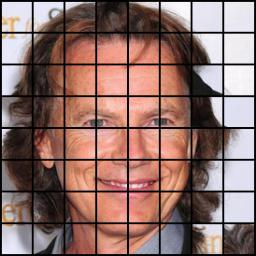}  & \includegraphics[width=0.110\textwidth]{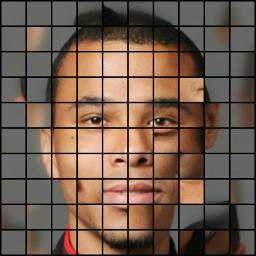} & \includegraphics[width=0.110\textwidth]{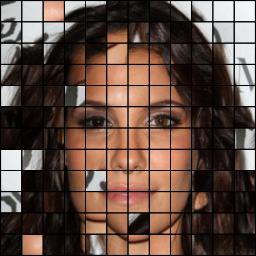} \\
    \bottomrule
    \end{tabularx}
    \centering
    \captionof{figure}{\label{fig:qual_celeb}Qualitative examples for PuzzleCelebA}
\end{minipage}
\hfill

\begin{minipage}{1.0\textwidth}
    \scriptsize
    \centering
    \begin{tabularx}{\textwidth}{X c c c c c c}
    \toprule
    \textbf{Model} & \textbf{2x2} &  \textbf{4x4}  & \textbf{6x6} & \textbf{8x8} & \textbf{10x10} & \textbf{12x12} \\
    \midrule

    Paikin and Tal \cite{PaikinCVPR2015} & - & - & \includegraphics[width=0.110\textwidth]{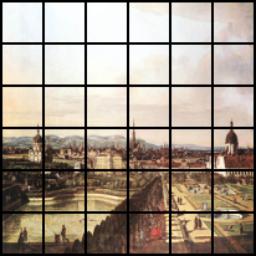}  & \includegraphics[width=0.110\textwidth]{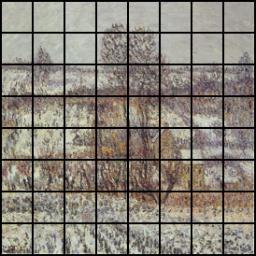}  & \includegraphics[width=0.110\textwidth]{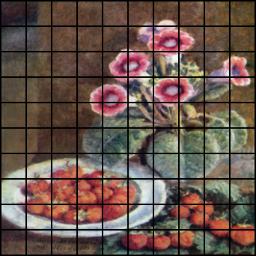} & \includegraphics[width=0.110\textwidth]{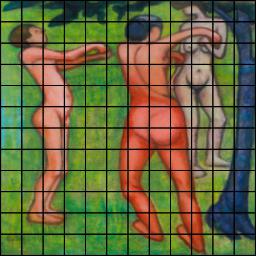} \\
    Pomeranz et al. \cite{pomeranzCVPR11greedy} & - & \includegraphics[width=0.110\textwidth]{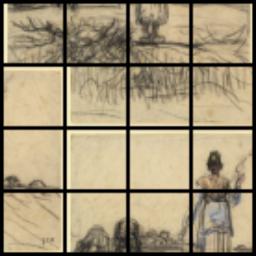} & \includegraphics[width=0.110\textwidth]{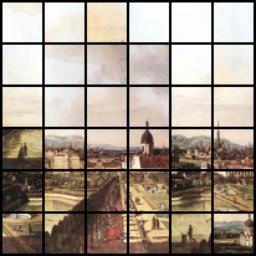}  & \includegraphics[width=0.110\textwidth]{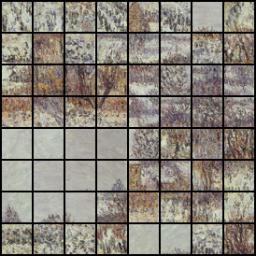}  & \includegraphics[width=0.110\textwidth]{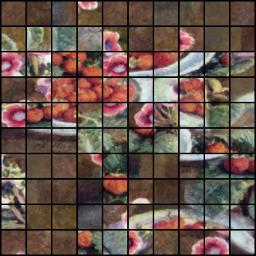} & \includegraphics[width=0.110\textwidth]{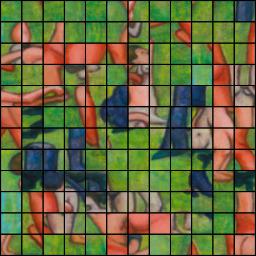} \\
    PO-LA \cite{zhang2018ICLR} & \includegraphics[width=0.110\textwidth]{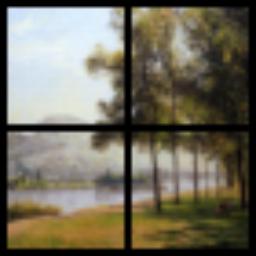} & \includegraphics[width=0.110\textwidth]{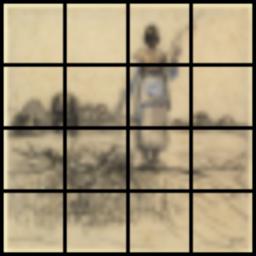} &\includegraphics[width=0.110\textwidth]{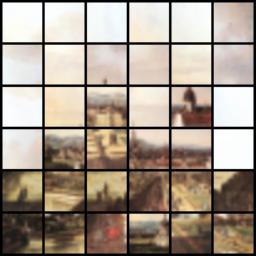} & \includegraphics[width=0.110\textwidth]{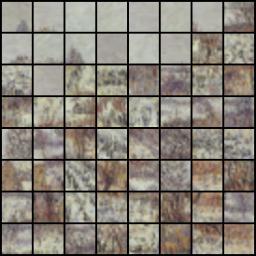}  & \includegraphics[width=0.110\textwidth]{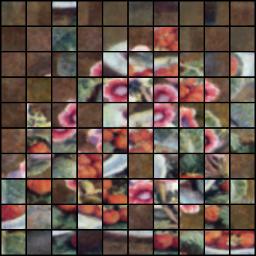}  & -\\
        Hung-perm & \includegraphics[width=0.110\textwidth]{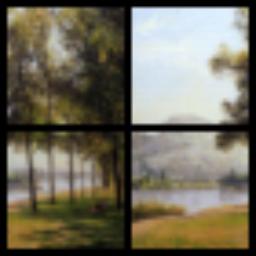} & \includegraphics[width=0.110\textwidth]{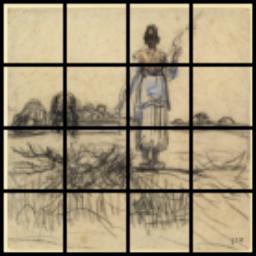}  & \includegraphics[width=0.110\textwidth]{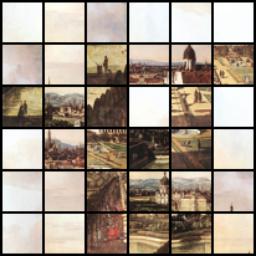}  & \includegraphics[width=0.110\textwidth]{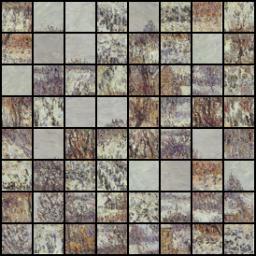} & \includegraphics[width=0.110\textwidth]{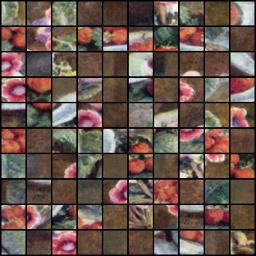} &
    \includegraphics[width=0.110\textwidth]{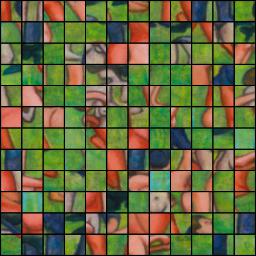} \\
    \midrule
    Ganzzle & \includegraphics[width=0.110\textwidth]{figures/results/ganzzle/ganzzle_wiki_2_2x2.jpg} & \includegraphics[width=0.110\textwidth]{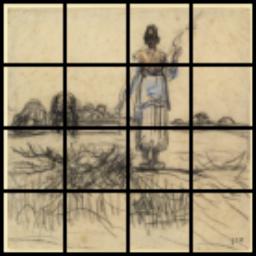} & \includegraphics[width=0.110\textwidth]{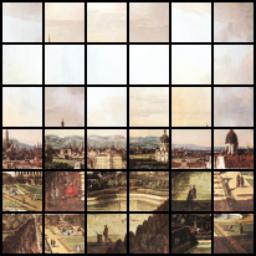}  & \includegraphics[width=0.110\textwidth]{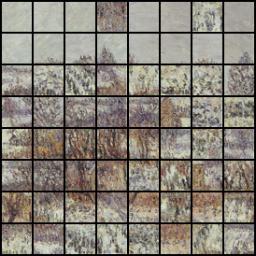}  & \includegraphics[width=0.110\textwidth]{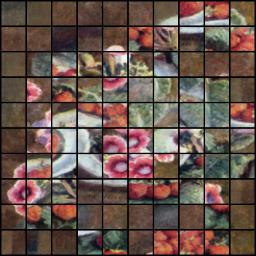} & \includegraphics[width=0.110\textwidth]{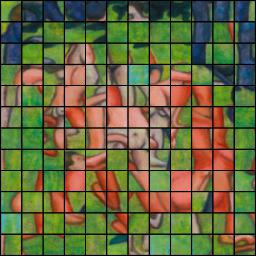} \\
    \bottomrule
    \end{tabularx}
    \centering
    \captionof{figure}{\label{fig:qual_wiki}Qualitative examples for PuzzleWikiArts}
\end{minipage}
\end{table*}

\end{document}